\documentclass{sn-jnl}
\usepackage{subcaption}

\jyear{2023}%

\theoremstyle{thmstyleone}%

\theoremstyle{thmstyletwo}%

\theoremstyle{thmstylethree}%

\begin{document}

\title[Post-Operative Complications Prediction]{Prediction of Post-Operative Renal and Pulmonary Complications Using Transformers}

\author[1]{\fnm{Reza} \sur{Shirkavand}}\email{res182@pitt.edu}

\author[2]{\fnm{Fei} \sur{Zhang}}\email{zhangfei@pitt.edu}

\author[1]{\fnm{Heng} \sur{Huang}}\email{heng.huang@pitt.edu}

\affil[1]{\orgdiv{Department of Electrical and Computer Engineering}, \orgname{University of Pittsburgh}, \orgaddress{\street{3700 O'Hara St}, \city{Pittsburgh}, \postcode{15213}, \state{PA}, \country{USA}}}

\affil[2]{\orgdiv{School of Nursing}, \orgname{University of Pittsburgh}, \orgaddress{\street{3500 Victoria Street}, \city{Pittsburgh}, \postcode{15261}, \state{PA}, \country{USA}}}

\abstract{ Postoperative complications pose a significant challenge in the healthcare industry, resulting in elevated healthcare expenses and prolonged hospital stays, and in rare instances, patient mortality. To improve patient outcomes and reduce healthcare costs, healthcare providers rely on various perioperative risk scores to guide clinical decisions and prioritize care. In recent years, machine learning techniques have shown promise in predicting postoperative complications and fatality, with deep learning models achieving remarkable success in healthcare applications. However, research on the application of deep learning models to intra-operative anesthesia management data is limited. In this paper, we evaluate the performance of transformer-based models in predicting postoperative acute renal failure, postoperative pulmonary complications, and postoperative in-hospital mortality. We compare our method's performance with state-of-the-art tabular data prediction models, including gradient boosting trees and sequential attention models, on a clinical dataset. Our results demonstrate that transformer-based models can achieve superior performance in predicting postoperative complications and outperform traditional machine learning models. This work highlights the potential of deep learning techniques, specifically transformer-based models, in revolutionizing the healthcare industry's approach to postoperative care.}

\maketitle

\section{Main}\label{sec1}
Post-operative complications present a significant challenge in the healthcare industry, resulting in elevated healthcare expenses, prolonged hospital stays, and, in rare instances \cite{pearse2006mortalityrate1,pearse2012mortalityrate2}, patient mortality. With approximately 230 million surgeries conducted globally each year \cite{weiser2008surgnum}, it is crucial for healthcare providers to predict potential complications and take preventive measures to improve patient outcomes and reduce healthcare costs.

Various perioperative risk scores are used to guide clinical decisions and prioritize care for patients undergoing surgery. Clinical risk scores help plan clinical management and allocate resources, while administrative risk scores help compare hospitals. The American Society of Anesthesiologists Physical Status Classification (ASA) \cite{dripps1963asa} and the Surgical Apgar \cite{gawande2007apgar} score are commonly used in the perioperative setting, but the latter has limitations such as subjectivity and variability. Logistic regression is used to develop new scoring systems \cite{le2016pospom}, and there are other recent perioperative administrative risk scores such as the Risk Stratification Index (RSI) \cite{sessler2010rsi} and the Risk Quantification Index (RQI) \cite{dalton2011rqi}.

Healthcare providers can use predictive models to identify high-risk patients and implement preemptive measures to prevent fatal post-operative complications. Machine learning techniques can be applied to analyze vast amounts of patient data, such as medical history, vital signs, laboratory values, and radiological findings, to identify patients at high risk of developing complications after surgery.


Deep Learning (DL) is a cutting-edge method for data classification and prediction, and is seen as the most promising solution to problems of data complexity. DL has achieved remarkable success in industry, as evidenced by recent studies \cite{nvidia2018}. In healthcare, DL has been used for computational phenotyping \cite{lasko2013computationalpheno,che2015deepcomppheno}, risk prediction \cite{cheng2016risk,miotto2016deep}, medical imaging analysis \cite{esteva2017dermatologist}, and natural language processing \cite{muneeb2015evaluating}. 

Although there have been significant achievements in deep learning models for other areas of healthcare, research on the application of deep learning models to intra-operative anesthesia management data is limited. This type of data has many advantages over typical EMR data(i.e. low-resolution AIMS data and high-resolution data directly from intra-operative physiological monitors), including data heterogeneity, longitudinal irregularity, inherent noise, and incomplete nature. However, the research in this area has mainly targeted static data such as demographic data, medical history, and post-operative complications \cite{muneeb2015evaluating,fritz2019deep}. In fact, there is a scarcity of research on the application of deep learning models to dynamic intra-operative physiological data.
In a real-time post-cardiosurgical care setting, \cite{meyer2018rnncomplication} utilized recurrent neural networks to predict severe complications like mortality, renal failure requiring renal replacement therapy, and postoperative bleeding leading to operative revision. In another study \cite{xue2021mlpulmonary}, logistic regression and decision tree variants were used to predict postoperative pulmonary complications. Similarly, \cite{lee2018mortalityprediction} employed feed-forward networks to predict post-operative mortality in critically ill patients.

Traditional machine learning methods, such as Gradient Boosting Decision Trees (GBDT) and its variants, including XGBoost \cite{chen2016xgboost}, LightGBM \cite{ke2017lightgbm}, and CatBoost \cite{prokhorenkova2018catboost}, have dominated the tabular data domain. Nonetheless, recent research has shown that deep learning (DL) methods can also achieve impressive results. For example, TabNet \cite{arik2021tabnet} introduced sequential attention to reason based on the most salient features, thereby providing interpretability. Neural Oblivious Decision Ensembles (NODE) \cite{popov2019node} incorporate hierarchical representation learning into the network. GrowNet \cite{badirli2020grownet} utilizes gradient boosting frameworks to construct a complex deep neural network from simple shallow components. Self Normalizing Neural Networks (SNN) \cite{klambauer2017snn} apply strong regularization and encourage zero mean and unit variance activations to facilitate high-level feature learning.

Transformers \cite{vaswani2017transformer}, initially introduced in the field of Natural Language Processing (NLP), are a type of neural network architecture specifically designed to handle sequential data. While they have been highly effective in NLP tasks, such as machine translation and text summarization, they have also been shown to be well-suited for tabular data analysis \cite{gorishniy2021rdtl}.

In this work, we aim to evaluate the performance of transformer-based models in predicting postoperative acute renal failure (PO-ARF), postoperative pulmonary complications (PPC), and postoperative in-hospital mortality from a tabular dataset of dynamic intra-operative anesthesia management variables. We compare the performance of our method with state-of-the-art prediction models suited to the dataset, including gradient boosting trees, and assess their ability to accurately predict the aforementioned complications using clinical data.

\section{Results}\label{sec2}
In this section, we present the results of our experiments comparing the performance of our transformer model to gradient boosting decision tree (GBDT) variants. The goal of our tasks were to predict postoperative pulmonary complications (PPC), the postoperative acute renal failure (PO-ARF), and postoperative in-hospital mortality.

Table \ref{tab:classification-results-complications} and Table \ref{tab:classification-results-mortality} show the accuracy, precision, recall, and F1 score of the models across 5-fold cross validation for both post-operative complication and post-operative in-hospital mortality prediction tasks. The transformer-based model outperforms all GBDT variants and deep baselines in all evaluation metrics.

\begin{table}[htbp]
\centering
\caption{Comparison of performance of baselines with the transformer model. Values correspond to the mean and standard deviation of each classification metric across 5-fold cross-validation of predicting post-operative pulmonary and renal complication}
\label{tab:classification-results-complications}
\resizebox{\textwidth}{!}{%
\begin{tabular}{@{}l|ll|ll|ll|ll@{}}
\toprule
\multirow{2}{*}{Method} & \multicolumn{2}{c|}{Accuracy} & \multicolumn{2}{c|}{Precision} & \multicolumn{2}{c|}{Recall} & \multicolumn{2}{c}{F1-score} \\ \cmidrule(l){2-9} 
                        & PPC             & PO-ARF      & PPC              & PO-ARF      & PPC            & PO-ARF     & PPC             & PO-ARF     \\ \midrule
SVM     & 0.803 ± 0.020 & 0.914 ± 003  & 0.779 ± 0.014  & 0.755 ± 0.007  & 0.802 ± 0.009  &  0.857 ± 0.017 & 0.785 ± 0.017   &  0.793 ± 0.008 \\ \midrule  
Logistic Regression     & 0.796 ± 0.024 &  0.896 ± 0.002 & 0.775 ± 0.015  & 0.730 ± 0.003  & 0.803 ± 0.011  & 0.882 ± 0.009  & 0.780 ± 0.020   & 0.777 ± 0.004  \\ \midrule 
Random Forest   & 0.813 ± 0.011 & 0.933 ± 0.007  & 0.807 ± 0.017  &  0.855 ± 0.025 & 0.748 ± 0.017  & 0.708 ± 0.033  & 0.765 ± 0.015   &  0.756 ± 0.034 \\ \midrule 
KNN   & 0.787 ± 0.015 & 0.916 ± 0.004  & 0.767 ± 0.022  & 0.824 ± 0.033  & 0.724 ± 0.014  & 0.596 ± 0.018  & 0.737 ± 0.016   & 0.634 ± 0.026  \\ \midrule 
XGboost                 &  0.817 ± 0.023     &   0.944 ± 0.009     & 0.797 ± 0.035       &   0.848 ± 0.025       & 0.751 ± 0.019      &     0.818 ± 0.030    &   0.766 ± 0.023    &   0.832  ± 0.028   \\ \midrule
LightGBM               &   0.822 ± 0.021    &    0.944 ± 0.009     &   0.794 ± 0.026   &   0.846 ± 0.023      &    0.775 ± 0.013    &   0.819 ± 0.036  &   0.782 ± 0.019       &    0.832 ± 0.031   \\ \midrule
Catboost                & 0.822 ± 0.022       &   0.945 ± 0.009    &    0.797 ± 0.028    &    0.849 ± 0.022     &  0.772 ± 0.014    &    0.819 ± 0.039     &    0.780 ± 0.019 &  0.833 ±  0.031 \\ \midrule
MLP                     & 0.831 ± 0.015      &   0.891 ± 0.006      & 0.798 ± 0.016       &    0.727 ± 0.010      & 0.807 ± 0.007     & 0.898 ± 0.004        & 0.802 ± 0.013       &   0.776 ± 010      \\ \midrule
SNN                      & 0.837 ± 0.002         &  0.900 ± 0.008     & 0.805 ± 0.003         &   0.737 ± 0.008     & 0.810 ± 0.003       &    0.894 ± 0.006   & 0.807 ± 0.003    &       0.785 ± 0.005     \\ \midrule
Node                    & 0.830 ± 0.003      &  0.922 ± 0.014      & 0.796 ± 0.004       &   0.778 ± 0.026       & 0.807 ± 0.005     &    0.879 ± 0.024     & 0.801± 0.004       &    0.814  ± 0.016    \\ \midrule
\textbf{Ours}  & \textbf{0.865 ± 0.005}      &   \textbf{0.967 ± 0.007}       & \textbf{0.830 ± 0.005}       &     \textbf{0.848 ± 0.009}     & \textbf{0.839 ± 0.004}     &      \textbf{0.917 ± 0.008}   & \textbf{0.834 ± 0.006}      &    \textbf{0.885 ± 0.007}     \\ \bottomrule
\end{tabular}
}
\end{table}

\begin{table}[htbp]
\centering
\caption{Comparison of performance of baselines with the transformer model. Values correspond to the mean and standard deviation of each classification metric across 5-fold cross-validation of predicting postoperative in-hospital mortality}
\label{tab:classification-results-mortality}
\resizebox{\textwidth}{!}{%
\begin{tabular}{@{}lllll@{}}
\toprule
Method & Accuracy & Precision & Recall & F1-score \\ \midrule
SVM     &   0.981 ± 0.001    & 0.748 ± 0.019      &  0.733 ± 0.018     & 0.740 ± 0.013  \\ \midrule
Logistic Regression     &   0.926 ± 0.008    &   0.597 ± 0.008    &    0.906 ± 0.006   &  0.641 ± 0.012 \\ \midrule
Random Forest     &  0.982 ± 0.001     &   0.875 ± 0.056    &  0.559 ± 0.004     &  0.598 ± 0.006 \\ \midrule
KNN     &    0.981 ± 0.001   &   0.804 ± 0.045    &  0.520 ± 0.005     &  0.533 ± 0.009 \\ \midrule
XGboost\cite{chen2016xgboost}                   & 0.990 ± 0.001                & 0.923 ± 0.010                & 0.781 ± 0.024              & 0.836 ± 0.019                \\ \midrule
Light GBM\cite{ke2017lightgbm}                  & 0.990 ± 0.001                & 0.909 ± 0.006                 & 0.789 ± 0.021              & 0.838 ± 0.016                \\ \midrule
Catboost\cite{prokhorenkova2018catboost}                   & 0.990 ± 0.001                & 0.919 ± 0.012                 & 0.792 ± 0.014             & 0.844 ± 0.009                \\ \midrule
MLP                        & 0.984 ± 0.002                & 0.771 ± 0.034                 & 0.856 ± 0.016              & 0.806 ± 0.017                \\ \midrule
SNN\cite{klambauer2017snn}                        & 0.983 ± 0.004                & 0.766 ± 0.041                 & 0.867 ± 0.029              & 0.804 ± 0.022                \\ \midrule
Node\cite{popov2019node}                       & 0.990 ± 0.001                & 0.881 ± 0.001                 & 0.800 ± 0.037              & 0.834± 0.022                 \\ \midrule
\textbf{Ours}              & \textbf{0.995 ± 0.005}                            &   \textbf{0.930 ± 0.006}                            &     \textbf{0.902 ± 0.004}                       &     \textbf{0.918 ± 0.005}                         \\ \bottomrule
\end{tabular}%
}
\end{table}

\begin{figure}[htbp]

\centering

\begin{subfigure}[b]{0.3\textheight}
\centering
\includegraphics[width=0.98\textwidth]{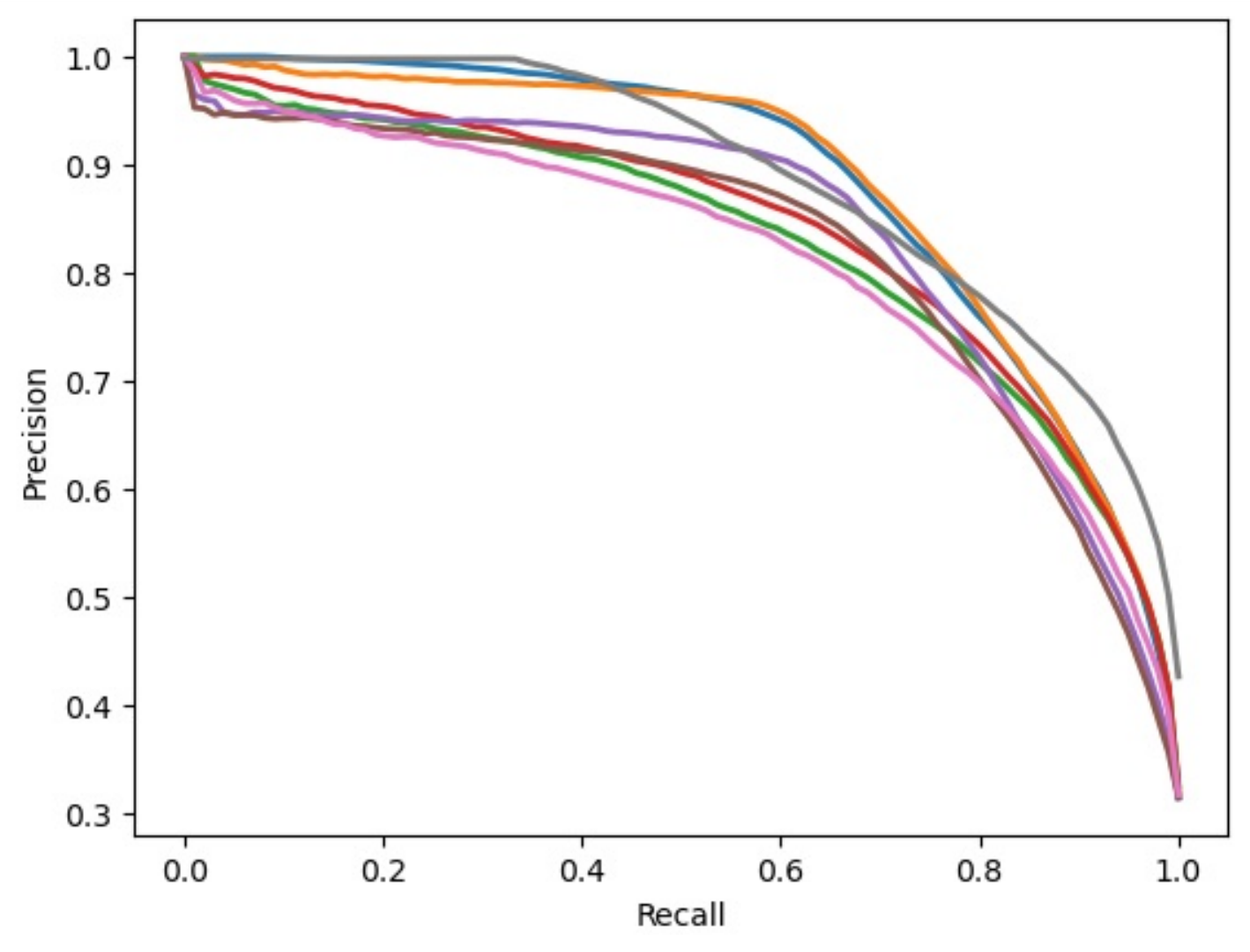}
\caption{Pulmonary complications}
\label{fig:prc-a}
\end{subfigure}
\\
\begin{subfigure}[b]{0.3\textheight}
\centering
\includegraphics[width=0.98\textwidth]{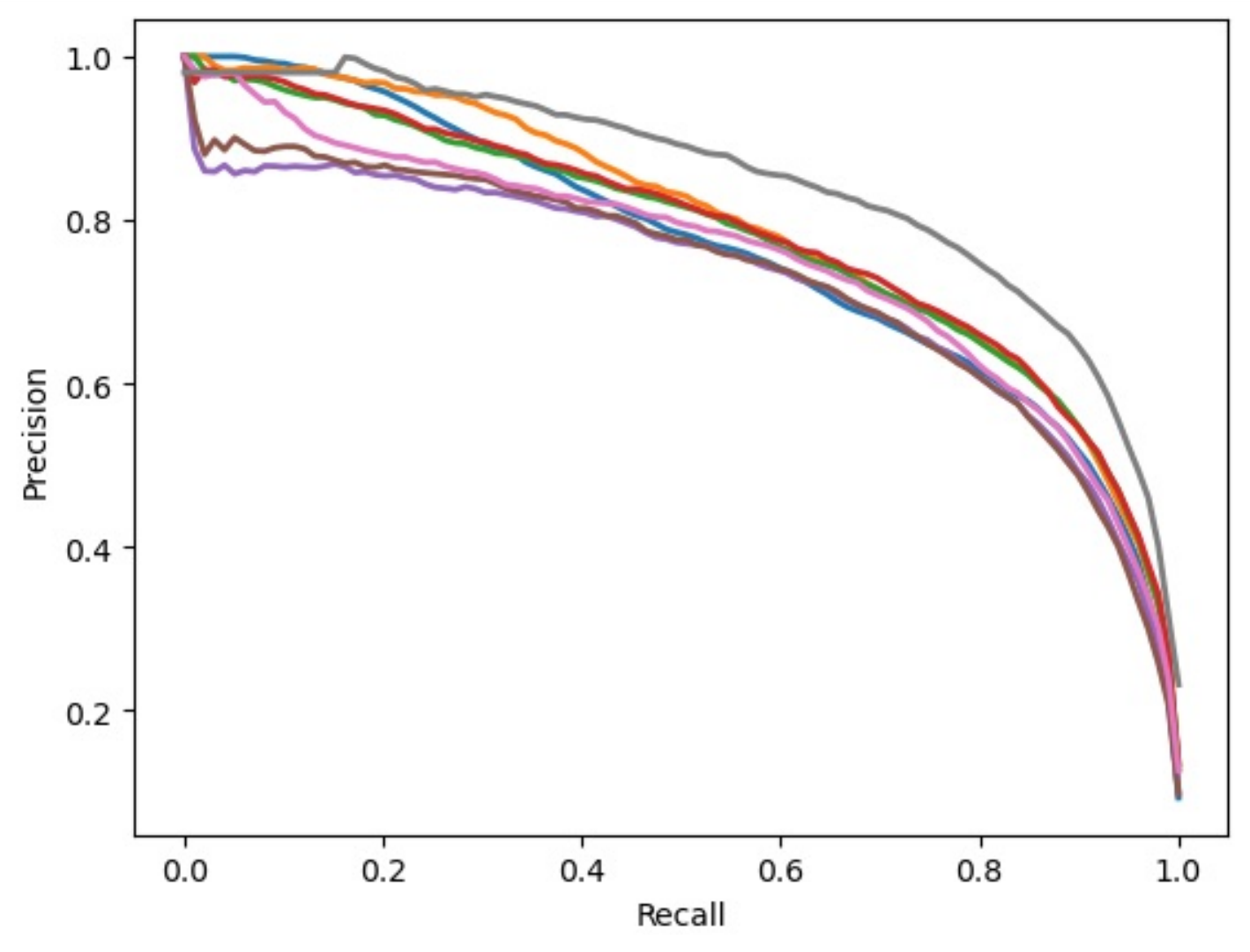}
\caption{Renal complications}
\label{fig:prc-b}
\end{subfigure}
\\
\begin{subfigure}[b]{0.3\textheight}
\centering
\includegraphics[width=0.98\textwidth]{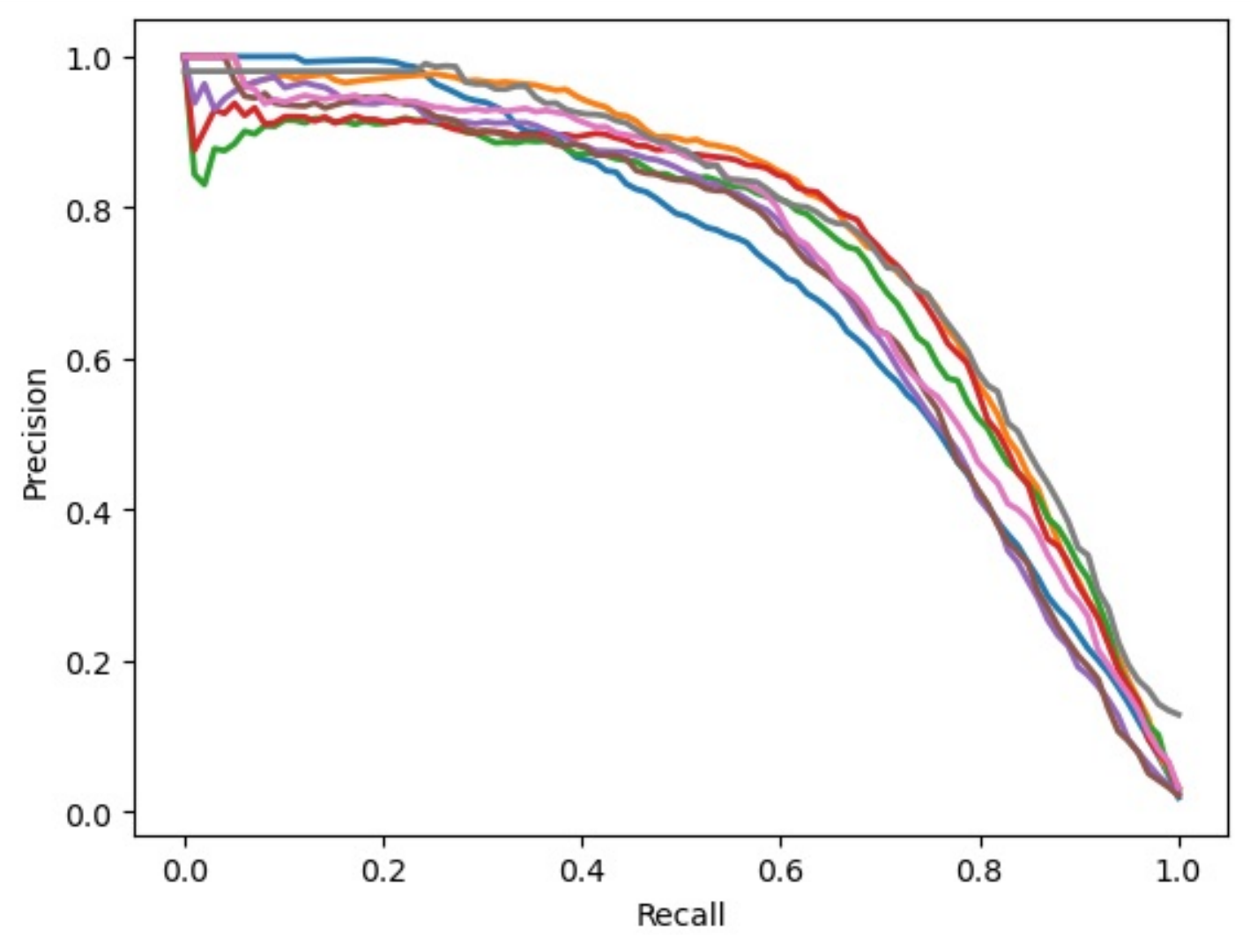}
\caption{In-hospital mortality}
\label{fig:prc-c}
\end{subfigure}

\caption{Precision Recall Curve of baselines as well as the transformer architecture for the prediction of pulmonary, renal complications, and in-hospital mortality}
\label{fig:prc}
\end{figure}

\begin{figure}[htbp]
\centering
\begin{subfigure}[b]{0.27\textheight}
\centering
\includegraphics[width=\textwidth]{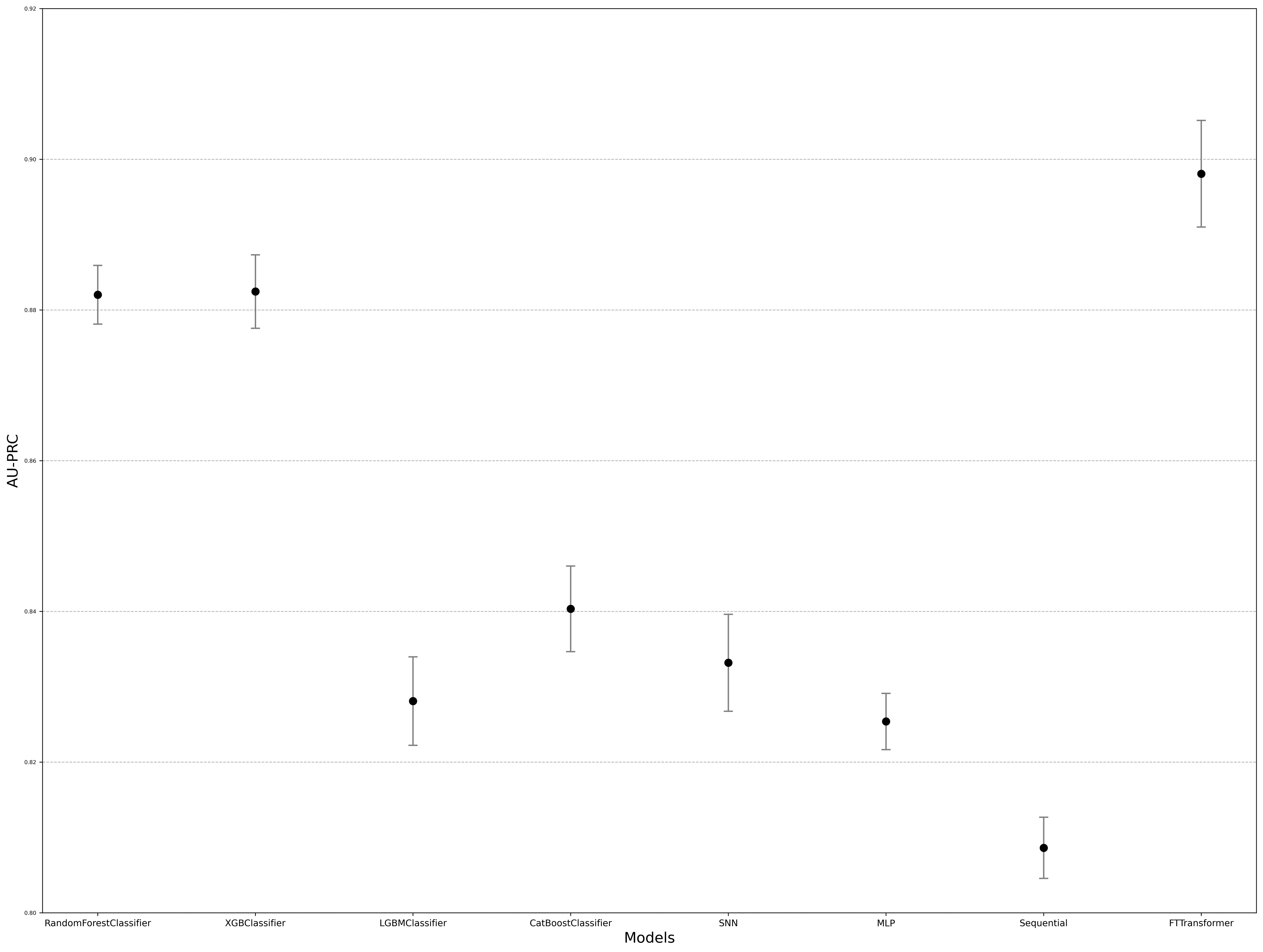}
\caption{Pulmonary complications}
\label{fig:auprc-a}
\end{subfigure}
\\
\begin{subfigure}[b]{0.27\textheight}
\centering
\includegraphics[width=\textwidth]{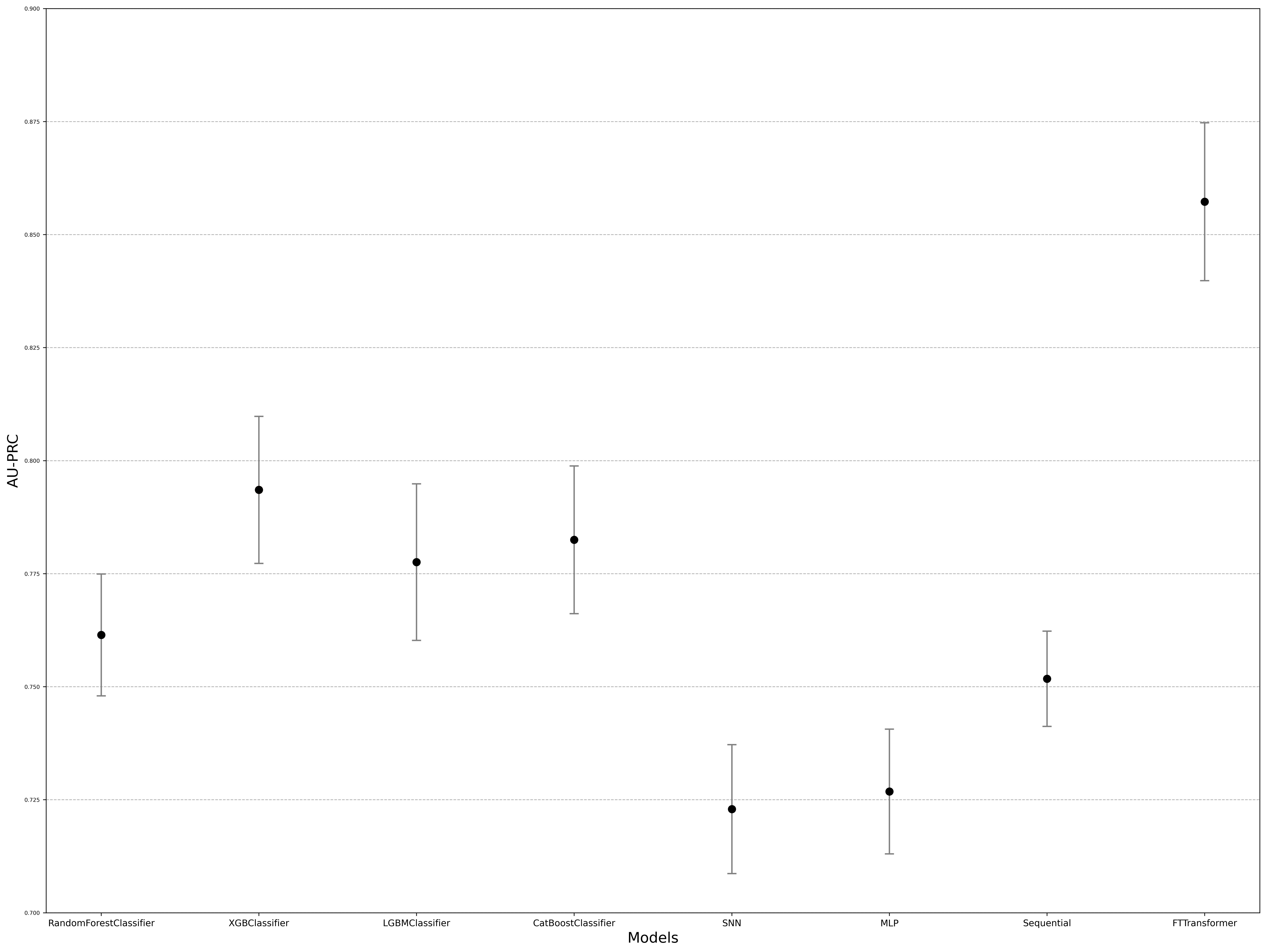}
\caption{Renal complications}
\label{fig:auprc-b}
\end{subfigure}
\\
\begin{subfigure}[b]{0.27\textheight}
\centering
\includegraphics[width=\textwidth]{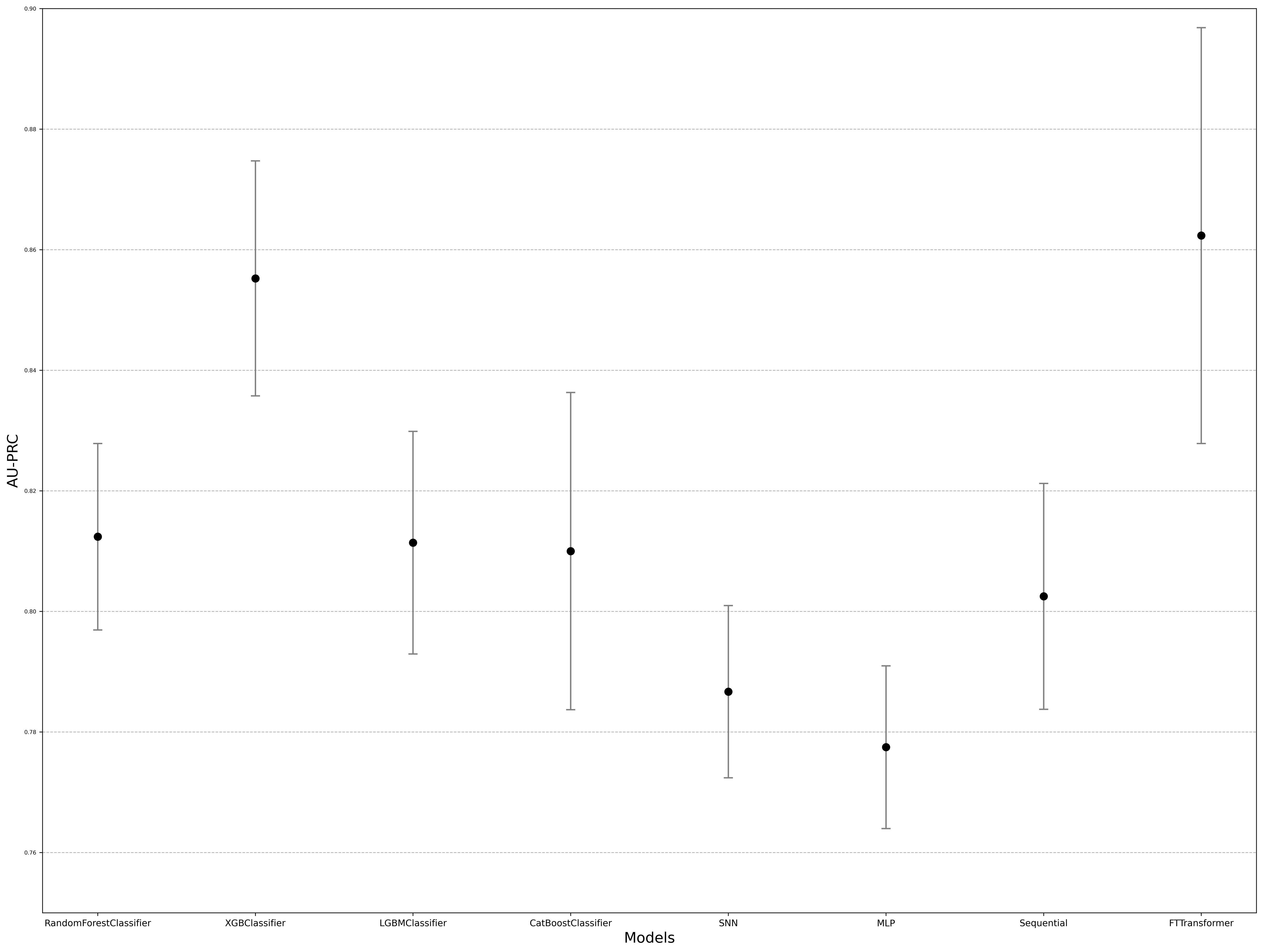}
\caption{In-hospital mortality}
\label{fig:auprc-c}
\end{subfigure}

\caption{Mean and standard deviation of area under precision recall curve (AUPRC) for all three task.}
\label{fig:auprc}
\end{figure}

We further analyzed the performance of the models by computing the precision recall curve.  Figure \ref{fig:prc} shows the Precision-Recall curve of all methods. The transformer-based model achieved the highest area under the curve (AUPRC), indicating that it has a better ability to accurately predict the complications. The error bars of AUPRC is displayed in Figure \ref{fig:auprc}

We attribute the substantial superiority of our method to several factors:
\begin{enumerate}
    \item Ability to learn complex relationships: Transformers are powerful models that are capable of learning complex relationships between variables. They can capture long-range dependencies in the data and model non-linear interactions between features, which may not be possible with simpler models like decision trees and MLPs.
    \item 
    Feature interactions: Tabular data often contains many features, and the interactions between these features can be crucial for making accurate predictions. Transformers are designed to handle feature interactions, and can learn to represent features in a way that captures their relationships with other features.
    \item 
    Robustness to missing data: Tabular data may have missing values (as was in our case where missing lab values where imputed with zeros), which can be a challenge for some models. Transformers can handle missing data by leveraging the attention mechanism to focus on relevant features and ignoring irrelevant ones.
\end{enumerate}
 It is worth noting that transformers have been shown \cite{gorishniy2021rdtl} to be a more universal model for structured data like our dataset.

\begin{figure}[htbp]
    \centering
    \includegraphics[width=\textwidth]{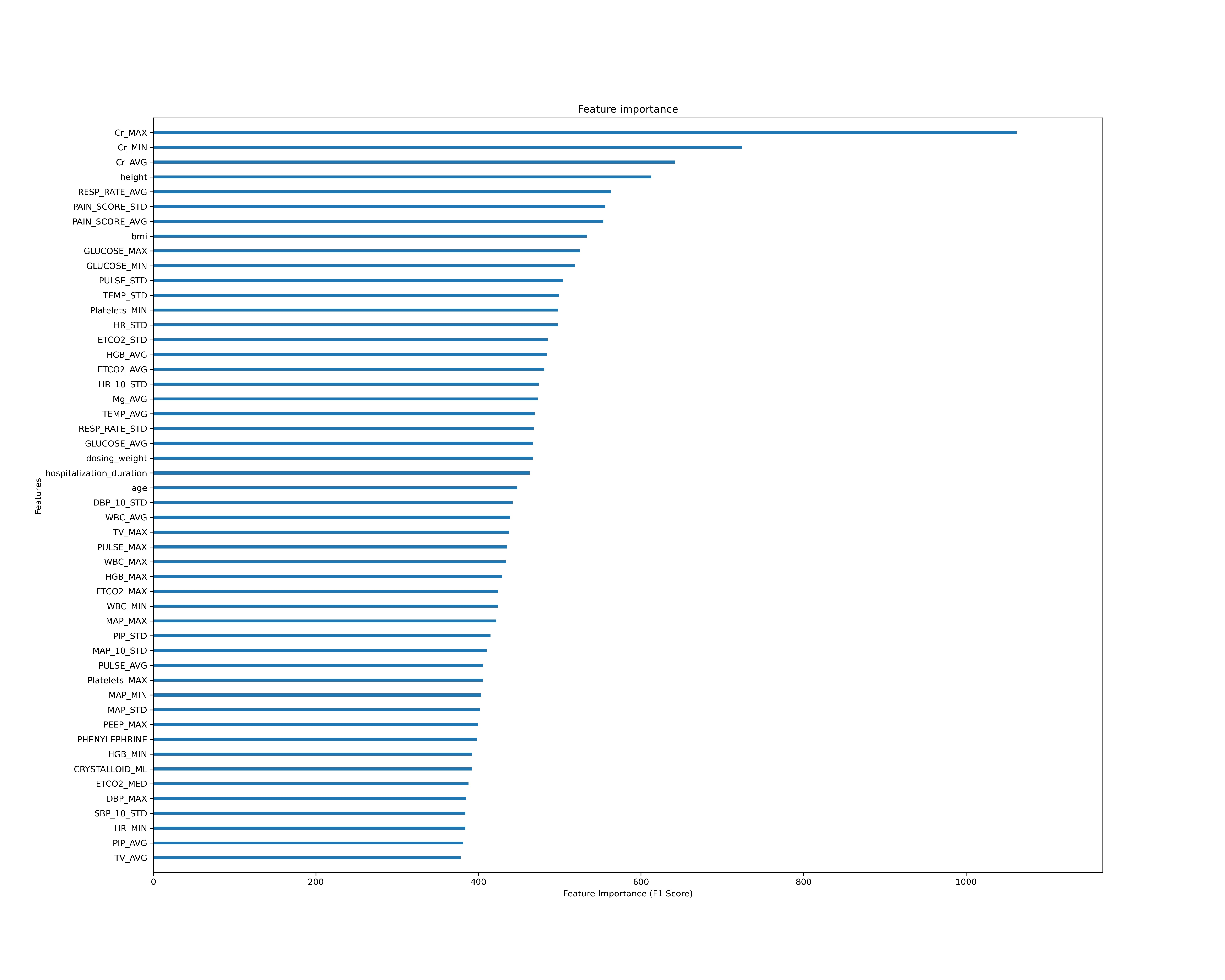}
    \caption{\textbf{Feature Importance Plot:} The graph shows the relative importance of features in predicting the target variable, as determined by permutation feature importance.}
    \label{fig:feature_importance}
\end{figure}

As shown in Figure \ref{fig:feature_importance}, the top 50 most important features of the dataset are plotted using a bar chart. The features were determined using permutation feature importance. Permutation feature importance works by randomly permuting the values of a single feature and measuring the decrease in model performance (e.g. F1 score) on a validation set. The larger the decrease in performance, the more important the feature is considered to be.

To calculate permutation feature importance for a transformer model, we:

\begin{enumerate}
    \item Train the transformer model on the dataset and evaluate the performance of the trained model on a separate validation set using the F1 score.
    \item 
    For each feature in the validation set, randomly permute its values and re-evaluate the F1 score. The difference between the original F1 score and the permuted F1 score is the feature importance score for that feature.
    \item 
    Repeat step 3 for all features in the validation set and rank them based on their importance scores.
\end{enumerate}

\section{Discussion}\label{sec3}
In this study, we conducted a retrospective analysis of the relationship between perioperative anesthesia management and patient outcomes for major surgeries under general anesthesia. We found that transformer-based models can effectively predict in-hospital morbidity and mortality based on automatically extractable and objective perioperative data.

Major surgery with general anesthesia can cause renal hypoperfusion, insufficient oxygen delivery, and eventually postoperative renal failure. Additionally, intraoperative neuromuscular blocking drugs and ventilation can change respiratory drive and muscle function, increasing the risk of postoperative pulmonary complications. These complications are associated with increased in-hospital mortality and are difficult to predict due to their heterogeneity. Risk factors for postoperative renal and pulmonary complications are numerous, thus anesthesia providers must recognize modifiable and non-modifiable risk factors and optimize patient care accordingly. Many validated risk prediction models are described. 

We compared the transformer-based model's performance to other validated risk prediction models, demonstrating that the transformer-based model outperforms other models due to its ability to capture inter-dependencies in the extracted variables important for predicting postoperative complications. Our findings have significant implications for improving postoperative care by providing simple and reliable models for individualized patient outcome monitoring and prediction, and learning capability. 

Our study has limitations, too. The study was conducted at a single institution, which may limit the generalizability of our findings. Future research should aim to validate the models' performance in larger and more diverse datasets across multiple institutions.

In conclusion, our study highlights the potential of  deep learning models, especially transformer-based models, in predicting postoperative complications and mortality, which could revolutionize the healthcare industry's approach to postoperative care by improving patient outcomes, reducing healthcare costs, and providing healthcare providers with tools to make informed clinical decisions.

\section{Methods}\label{sec4}
\subsection{Data Extraction}

The Department of Anesthesiology and Perioperative Medicine at the University of Pittsburgh (Pitt) and University of Pittsburgh Medical Center (UPMC) is widely considered to be the largest academic anesthesiology program in the United States. UPMC has more than 300 surgical sites that are equipped with Anesthesia Information Management System (AIMS) since 2012. UPMC uses Cerner Powerchart  solution as its EMR system and Surginet as its AIMS solution. The necessary raw clinical data were retrieved from the UPMC clinical database by the UPMC Health Record Research Request team based on the research team's data retrieval strategy. We are interested in the major surgeries that may involve hemodynamic instability and require invasive hemodynamic monitoring. Totally perioperative clinical data from 67,295 perioperative cases with155 different types of surgeries are collected for further data analysis.
Given the array of metrics needed and the complexity of the data structure in the Cerner database, a 2-stage data warehouse is constructed to reduce the need to join and optimize multiple tables (e.g., variables and files).The first stage, termed "Base Tables", is designed to serve as a middle layer decreasing the number of tables and simplifying the joins between them. Conceptually, the tables in Cerner's Millennium coalesce into 3 groups: (1) patient-centered variables (laboratories, allergies, medical history, etc.); (2) encounter centered files (admission notes, discharge notes, pre-op anesthesia evaluation note, etc.); and (3) patient-encounter hybrid tables (Anesthesia Record, etc.). Tables are joined by the following fields: (1) a patient identifier (e.g., name, medical number, fin number) in Powerchart and SurgiNet; and (2) a case identifier (e.g., location, name of surgery, case number) in Cerner SurgiNet.

\subsection{Target Definition}

To identify the presence of a specific postoperative complication, the study relied on the documentation of a corresponding complication code in the EMR. The "present\_on\_admit" field in the EMR was used to indicate whether the complication was present on admission, and if it was marked as "N" or left empty, it suggested that the complication developed after operation.

To identify in-hospital mortality, the study used a binary event [0, 1]. The event was considered present if a related note, such as a death summary, death pronouncement note, or death certificate, was found in the patient's record associated with the death. Importantly, the definition of in-hospital mortality was not dependent on the patient's length of stay in the hospital.

\subsection{Extracted Input Features}
For each hospital admission, a surgical record was created that contained 151 distinct features, which were computed or extracted at the conclusion of the surgery. The features consisted of various intra-operative vital signs, such as minimum and maximum blood pressure values, and a summary of drugs and fluids interventions, including the total amount of blood infused and Vasopressin administered. Patient anesthesia descriptions were also included, such as the presence of an arterial line and the type of anesthesia utilized. To review the complete list of features, refer to Table 


\subsection{Transformers}

Transformers are a type of neural network architecture used for a wide range of natural language processing (NLP) tasks such as machine translation, sentiment analysis, and language modeling. They were introduced in the seminal paper \cite{vaswani2017transformer} as a replacement for recurrent neural networks (RNNs) and have since become the state-of-the-art method for many NLP tasks.

At the core of the transformer architecture is the concept of self-attention, which allows the model to weigh the importance of different tokens in a sequence when making predictions. Self-attention is a type of attention mechanism that can be computed in parallel across all tokens in a sequence. The transformer architecture uses a multi-head self-attention mechanism that is capable of attending to multiple positions at once.

The self-attention mechanism is composed of several layers, each of which has three main components: the query, the key, and the value. These components are derived from the input sequence, and are used to calculate the attention weights.

The query, key, and value components are obtained by applying learned linear transformations to the input sequence. Specifically, given an input sequence $\mathbf{X} = [\mathbf{x}_1, \mathbf{x}_2, \dots, \mathbf{x}_n]$, the query, key, and value components are calculated as follows:

\begin{equation}
\mathbf{Q} = \mathbf{X} \mathbf{W}_Q,
\mathbf{K} = \mathbf{X} \mathbf{W}_K,
\mathbf{V} = \mathbf{X} \mathbf{W}_V
\end{equation}
where $\mathbf{W}_Q$, $\mathbf{W}_K$, and $\mathbf{W}_V$ are learned weight matrices.

The attention weights are then calculated using the dot product between the query and key vectors, scaled by the square root of the dimension of the key vectors:

\begin{equation}
    \mathrm{Attention}(\mathbf{Q}, \mathbf{K}, \mathbf{V}) = \mathrm{softmax}\left(\frac{\mathbf{Q} \mathbf{K}^\top}{\sqrt{d_k}}\right) \mathbf{V}
\end{equation}

where $d_k$ is the dimension of the key vectors. The softmax function is applied row-wise to ensure that the attention weights for each position in the sequence sum to one.

Multi-head self-attention is an extension of self-attention that allows the model to attend to information from multiple representation subspaces. In other words, instead of computing a single set of query, key, and value vectors for each position in the sequence, multi-head attention computes multiple sets of vectors, each of which attends to a different subspace of the input representation. These multiple attention heads allow the model to capture more nuanced relationships between different parts of the input sequence and to learn more complex interactions between them. The multi-head self-attention result is obtained by concatenating the results of multiple attention heads:

\begin{equation}
    \mathrm{MultiHead}(\mathbf{Q}, \mathbf{K}, \mathbf{V}) = \mathrm{Concat}(\mathrm{head}_1, \mathrm{head}_2, \dots, \mathrm{head}_k) \mathbf{W}_O 
\end{equation}

\begin{equation}
    \mathrm{head}_i = \mathrm{Attention}(\mathbf{Q} \mathbf{W}_i^Q, \mathbf{K} \mathbf{W}_i^K, \mathbf{V} \mathbf{W}_i^V)
\end{equation}

 and $W_{i}^Q$, $W_{i}^K$, $W_{i}^V$, and $W_O$ are trainable weight matrices. The output of the multi-head self-attention layer is then passed through a feed-forward network with a non-linear activation function.

Our architecture is based on the Transformer design proposed in \cite{gorishniy2021rdtl}. The model takes the input $x$ and maps it to an embedding space $h$ using an affine transformation given by:
\begin{equation}
h_{j} = x_{j}.W_{j} + b_{j}
\end{equation}
Here, $x_{j}$ is the $j$-th feature of the data point $x$, and $W_{j}$ and $b_{j} \in R^{d}$ are the corresponding weight and bias vectors. The final embedding $h$ is obtained by concatenating all the embeddings of the input data:
\begin{equation}
h = Concat[h_{1}, \cdots , h_{d}],.
\end{equation}

The embeddings $h$ are then processed by $L$ Transformer blocks $f_{\theta}^{k}$ to generate the final representation $z$, which is given by:
\begin{equation}
z = f_{\theta}^{L} (\cdots f_{\theta}^{1}(h)),.
\end{equation}

\begin{figure}
    \centering
    \includegraphics[width=\textwidth]{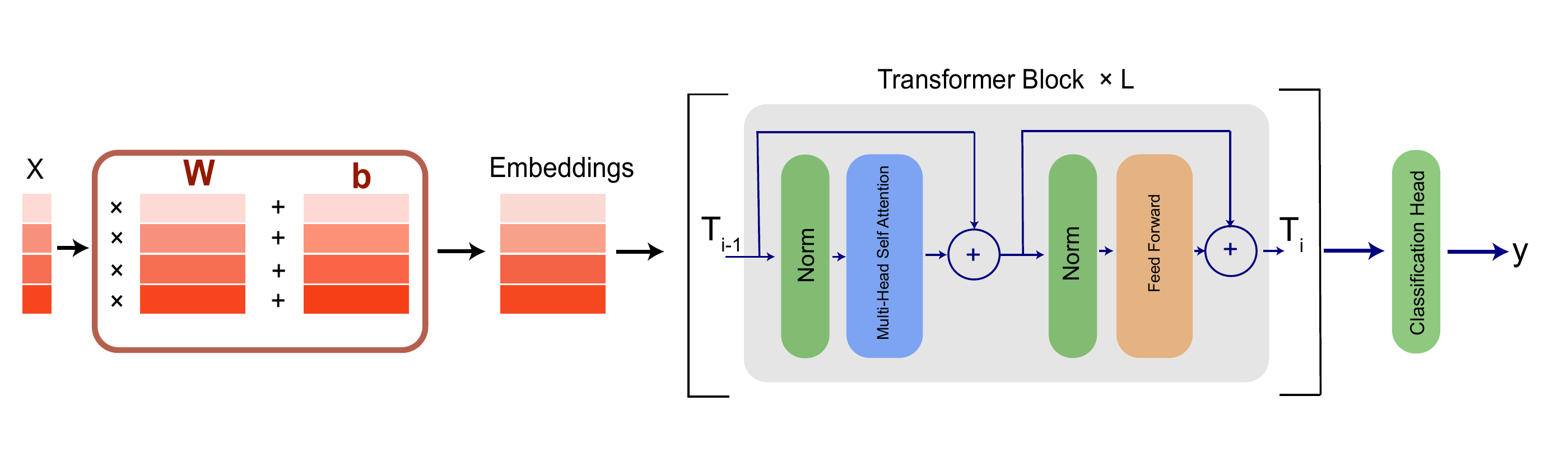}
    \caption{Overview of the transformer-based framework. First, 1-D features are mapped to an embedding space by an affine transformation. Then several transformer blocks are applied to produce the final output.}
    \label{fig:method}
\end{figure}

The final prediction $\hat{y}$ is obtained by passing a classification head $g$, as follows:
\begin{equation}
\hat{y} = g(z)
\end{equation}
Figure \ref{fig:method} depicts our transformer architecture.

\subsection{Baselines}
We conduct a comparative analysis of our framework and previous research on tabular data prediction. Our comparison is primarily focused on variants of Gradient Boosting Decision Trees (GBDT)\cite{friedman2001gbdt}, such as XGBoost \cite{chen2016xgboost}, LightGBM \cite{ke2017lightgbm}, and CatBoost \cite{prokhorenkova2018catboost}, which are widely used. However, we also examine recently proposed deep learning methods, including Neural Oblivious Decision Ensembles (NODE) \cite{popov2019node}, GrowNet \cite{badirli2020grownet}, Self-Normalizing Networks (SNN) \cite{klambauer2017snn}, as well as simple Multi-Layer Perceptron (MLP).

\subsection{Experimental Setup}

During the training process, we used the stratified 5-fold cross-validation method, which is an important technique for ensuring that the model is not biased towards any particular class. In stratified cross-validation, the dataset is split into k folds, where each fold contains roughly the same proportion of each class as the original dataset. This helps to ensure that the model is trained on a representative sample of the data and can generalize well to unseen data. The use of this technique helps to prevent overfitting and ensure that the model can accurately classify all classes, regardless of their frequency in the dataset.

We used the Binary Cross Entropy Loss function as the loss function during training. To ensure that the model learns to classify all classes accurately, we used the class-balanced version of the BCE function. This assigns higher weights to underrepresented classes and lower weights to overrepresented classes, effectively balancing the contribution of each class to the overall loss. The use of this function helps to prevent the model from ignoring the minority classes and ensures that all classes are learned equally. We used the AdamW optimizer with learning rate (lr) of 0.0003, betas of (0.9,0.999), and weight decay of 0.001.

We used a batch size of 256 and trained the model for 200 epochs. A large batch size helps to reduce the noise in the gradient estimation and can lead to faster convergence. We also used early stopping with a patience of 10, which means that the training was stopped if the validation loss did not improve for 10 consecutive epochs. This helped to prevent overfitting and ensure that the model was not trained for too long, which can lead to poor generalization performance.
The value of all hyperparameters were chosen based on previous experiments and empirical observations

We trained the baselines using a set of carefully chosen hyperparameters selected from a pool of possible values. Our aim was to ensure that the models could achieve the highest possible performance on the given tasks. The training settings employed to train the baselines can be found in the publicly available GitHub repository of our paper. This repository comprises all the essential code, scripts, and instructions required for replicating our experiments and the results reported in the paper. Our goal in providing this information is to facilitate other researchers' ability to expand on our work and explore new avenues of research.

\bibliographystyle{bst/sn-standardnature.bst}
\bibliography{references}

\end{document}